\documentclass{article}

\usepackage{arxiv}

\usepackage[utf8]{inputenc} 
\usepackage[T1]{fontenc}    
\usepackage{hyperref}       
\usepackage{url}            
\usepackage{booktabs}       
\usepackage{amsfonts}       
\usepackage{nicefrac}       
\usepackage{microtype}      
\usepackage{lipsum}
\usepackage{amsmath}
\usepackage{algorithm}
\usepackage{algorithmic}
\usepackage{graphicx}
\usepackage[sorting=none]{biblatex}
\usepackage{multicol}
\title{Learning Equational Theorem Proving}

\author{
  Jelle Piepenbrock \thanks{Radboud University Nijmegen, The Netherlands} \ \footnotemark[2] \\
  
   \And
   Tom Heskes \footnotemark[1]
   \And Mikol\'a\v{s} Janota \thanks{Czech Technical University Prague, Czech Republic}
   \And Josef Urban \footnotemark[2]
}

\begin{document}
\maketitle

\begin{abstract}
We develop Stratified Shortest Solution Imitation Learning (3SIL) to
learn equational theorem proving in a deep reinforcement learning (RL)
setting. The self-trained models achieve 
state-of-the-art performance in proving problems generated by one of
the top open conjectures in quasigroup theory, the Abelian Inner
Mapping (AIM) Conjecture. To develop the methods, we first use two simpler arithmetic rewriting tasks that share tree-structured proof states and sparse rewards with the AIM problems. On these tasks, 3SIL is shown to significantly outperform several established RL and imitation learning methods. The final system is then evaluated in a standalone and cooperative mode on the AIM problems. The standalone 3SIL-trained system proves in 60
seconds more theorems (70.2\%) than the complex, hand-engineered
\textit{Waldmeister} system (65.5\%). In the cooperative mode, the final system
is combined with the \textit{Prover9} system, proving in 2 seconds
what standalone \textit{Prover9} proves in 60 seconds.

\end{abstract}


\section{Introduction}
Machine learning (ML) has recently proven its worth in many separate fields. From computer vision \cite{he2016deep}, to speech recognition \cite{graves2006connectionist} to game playing \cite{mnih2015human, silver2016mastering, sutton2018reinforcement} using \emph{reinforcement learning} (RL), machine learning has made immense progress. In automated mathematics and automated theorem proving, machine learning has also recently achieved some encouraging results. 

Learned models have been used in saturation-based theorem provers to improve given clause selection  \cite{chvalovsky2019enigma}, to synthesize functions in higher-order logic \cite{gauthier2020deep} and to guide a connection-tableau prover using reinforcement learning \cite{kaliszyk2018reinforcement}. There have also been efforts to use reinforcement learning in interactive theorem proving \cite{bansal2019holist, gauthier2020tactictoe}.  

In general, the knowledge base of mathematics and the complexity of mathematical proofs will grow in the future. The need for proof checking and higher computer mathematics performance will therefore increase. Simultaneously, the complexity of software and the need for formal verification to prevent failure modes have also been growing \cite{de2008z3}. Automated theorem proving and mathematics will benefit from advanced ML integration. One of the mathematical subfields where automated theorem provers are heavily used is the field of quasigroup and loop theory \cite{phillips2010automated}. A \textit{quasigroup} is similar to a group, but it does not guarantee associativity. A \textit{loop} is a quasigroup with an identity element.

In this paper, we present Stratified Shortest Solution Imitation Learning (3SIL), a new machine learning approach that can learn equational theorem proving. We compare our 3SIL algorithm on two simpler arithmetic tasks with established reinforcement learning baselines and then apply it on theorems generated in the context of the \textit{Abelian Inner Mapping (AIM) Conjecture} for loops \cite{kinyon2013loops} using the AIMLEAP proof environment \cite{brown2020learning}. 

The structure of the paper is as follows. Sections \ref{section:ar} and \ref{section:aim} describe the datasets used for the arithmetic and AIM conjecture tasks respectively. The RL methods used as baselines and the proposed 3SIL method are discussed in Section \ref{section:training}. Afterwards, Section \ref{section:models} describes the tree neural network models used to process the mathematical expressions. Section \ref{section:ar_exp} focuses on the experiments performed on the arithmetic tasks and comparison with RL methods. Our results on the AIM conjecture dataset and comparison with state-of-the-art automated theorem provers are in Section \ref{section:aim_exp}. Discussion and conclusions are in Sections \ref{section:disc} and \ref{section:conc}. Our contributions are the following.

\begin{enumerate}
    \item We release a Python reinforcement learning environment for arithmetic rewriting tasks.
    \item We propose 3SIL, stratified shortest solution imitation learning, a method that outperforms established reinforcement learning methods on the arithmetic rewriting tasks in our environments.
    \item We use 3SIL to learn equational theorem proving in the AIMLEAP proof environment. The system performs competitively with established equational theorem provers, that contain many hours of work and lines of code.  This is shown on a significant mathematical domain: the theory surrounding the open AIM conjecture, one of the top open problems in quasigroup theory.
    \item We combine the learned system with an existing state-of-the-art automated theorem prover. The combined system outperforms the original prover, even with a smaller time limit.
\end{enumerate}

\section{Arithmetic Tasks}
\label{section:ar}
To develop a method that can solve equational theorem proving problems, we considered two simpler arithmetic tasks, which also have a tree-structured input and a sparse reward structure: Robinson arithmetic and polynomial arithmetic. In both cases, the task is to normalize a mathematical expression to one specific form. These  two tasks are implemented as Python RL environments, which we make available.

The learning environments incorporate two existing datasets. For the Robinson arithmetic normalization task, we use a dataset that was constructed for reinforcement learning experiments in the interactive theorem prover HOL4 \cite{gauthier2019deep}. For the polynomial normalization task, we employ a dataset introduced for experiments in symbolic rewriting using recurrent language models  \cite{piotrowski2019can}. 
\subsection{Normalization in Robinson Arithmetic}
The first formalism that we use as an RL environment is Robinson arithmetic (RA). RA is a simple arithmetic theory. Its language contains the successor function \textit{S}, addition \textit{+} and multiplication \textit{*} and one constant, the 0. The theory considers only non-negative numbers. Numbers are represented by the constant 0 with the appropriate number of successor functions applied to it. The task for the agent is to rewrite an expression until there are only nodes of the successor or 0 types. Effectively, we are asking the agent to calculate the value of the expression. As an example, $S(S(0)) + S(0)$, representing $2+1$, needs to be rewritten to $S(S(S(0)))$. The setup for this RA normalization task is modeled after \cite{gauthier2019deep}.

The expressions are represented as a tree data structure.
Within the environment, there are seven different rewrite actions available to the agent.  The four equations defining these actions are $x + 0 = x$, $ x + S(y) = S(x + y)$, $ x * 0 = 0$ and $x * S(y) = (x * y) + x$, where the agent can apply the equations in either direction. There is one exception: the multiplication by 0 cannot be applied from right to left, as this would require the agent to supply a term that would function as the \textit{x}. The place where the rewrite is applied is denoted by the location of a \textit{cursor} in the expression tree.

In addition to the rewrite actions, the agent can move the cursor to one of the children of the current cursor node. This gives a total number of nine actions. Moving to a child of a node with only one child counts as moving to the left child. After a rewriting action, the cursor is reset to the root of the expression. More details on the actions are in Appendix 1.
\subsection{Normalization in Polynomial Arithmetic}
The polynomial normalization task can be seen as a more challenging extension of the RA normalization task. The polynomial arithmetic implemented follows the axioms known as \textit{Tarski's high school axioms} \cite{burris1993tarski}. These introduce exponentiation, distributivity and associativity and free variables (for example \textit{x, y, z}). We still represent numbers as successor functions applied to the constant 0. An example of a problem is the normalization of $( ( z * y ) ^ 3 ) * ( ( ( 2 * 0 ) + y ) * ( ( 1 * x ) + z ) )$ to $xy ^ 4 z ^ 3 + y ^ 4  z ^ 4.$ Due to the larger number of axioms in this arithmetic, there are more rewriting actions. In total, the polynomial arithmetic environment allows the agent to perform 28 different actions, including two actions that involve cursor movement (see Appendix 2 for details). 

Both RL environments incorporate all the necessary logic to rewrite the tree correctly. The agents only supply an action to change the tree or to move the cursor to another node in the tree. Both normalization tasks only give a reward at the very end, when the expression is normalized. This reward structure is generally difficult to deal with, as there is no indication for the agent to know that it is on the right track, or to come up with partial solutions. It is also the reward structure for many theorem proving settings. For these normalization problems, there is an efficient algorithm that is guaranteed to correctly rewrite every expression, but this algorithm is not necessarily trivial to learn.

\section{AIM Conjecture Task}
\label{section:aim}

Automated theorem proving has been applied in the theory surrounding the Abelian Inner Mapping Conjecture, known as the AIM Conjecture. This is one of the top open conjectures in quasigroup theory. Work on the conjecture has been going on for more than a decade. Automated theorem provers use hundreds of thousands of inference steps when run on problems from this theory. Theorem proving is undecidable, meaning that there cannot possible be a computer program that always gives the correct answer.

There is a dataset of theorems generated by this conjecture, which we use as a testbed for our machine learning methods \cite{brown2020learning}. The dataset comes with a simple prover called AIMLEAP that can take machine learning advice.\footnote{\url{https://github.com/ai4reason/aimleap}} We use this system as a reinforcement learning environment. AIMLEAP keeps the state and carries out the cursor movements and rewrites that a model chooses. 

The AIM conjecture concerns specific structures in \textit{loop theory} \cite{kinyon2013loops}. A loop is a quasigroup with an identity element. A quasigroup is a generalization of a group that does not preserve associativity. This mostly manifests in the presence of two different `division' operators, one left-division and one right-division.  We will briefly explain the conjecture because it indicates the nature of the data.

For loops, three \textit{inner mapping functions} (left-translation L, right-translation R, and the mapping T) are:
\begin{align*}
    L(u, x, y) &:= (y * x) \backslash (y * (x * u)) \\
    R(u, x, y) &:= ((u * x) * y) / (x * y) \\
    T(u, x) &:= x \backslash (u * x).
\end{align*}
These mappings can be seen as measures of the deviation from commutativity and associativity. The conjecture concerns the consequences of these three inner mapping functions forming an Abelian (commutative) group. There are two additional notions, that of the \textit{associator} function \textit{a} and the \textit{commutator} function \textit{K}:
\begin{align*}
    a(x,y,z) &:= (x * (y * z)) \backslash ((x * y) * z) \\
    K(x,y) &:= (y * x) / (x * y).
\end{align*}
From these definitions, the conjecture can be stated. There are two parts to the conjecture. For both parts, the following equalities need to hold for all \emph{u, v, x, y,} and \emph{z}:
\begin{align*}
   a(a(x,y,z),u,v) &= 1 \\
   a(x,a(y,z,u),v) &= 1 \\
   a(x,y,a(z,u,v)) &= 1 
\end{align*}
where 1 is the identity element. These are necessary, but not sufficient for the two main parts of the conjecture. The first part of the conjecture asks whether a loop modulo its center is a group. In this context, the \textit{center} is the set of all elements that commute with all other elements. This is the case if:
\begin{align*}
    K(a(x,y,z),u) &= 1. 
\end{align*}
The second part of the conjecture asks whether a loop modulo its nucleus is an Abelian group. The \textit{nucleus} is the set of elements that associate with all other elements. This is the case if:
\begin{align*}
    a(K(x,y),z,u) &= 1  \\   
    a(x,K(y,z),u) &= 1  \\  
    a(x,y,K(z,u)) &= 1. 
\end{align*}
Currently, work in this area is done using automated theorem provers such as \textit{Prover9} \cite{prover9-mace4, kinyon2013loops}. This has led to some promising results, but the search space is enormous. The main strategy for proving the AIM conjecture thus far has been to prove weaker versions of the conjecture (using additional assumptions) and then import crucial proof steps into the stronger version of the proof. The \textit{Prover9} theorem prover is especially suited to this approach because of its well-established \textit{hints} mechanism \cite{DBLP:journals/jar/Veroff96}. The dataset is derived from this \textit{Prover9} approach and contains around 3500 theorems that can be proven with the supplied definitions and lemmas \cite{brown2020learning}.

 There are 177 possible actions in the environment. Three actions are cursor movements, where the cursor can be moved to an argument of the current position. The other actions all rewrite the current term at the cursor position with various axioms, definitions and lemmas that hold in the AIM context (see Appendix 3 for a list of these actions).

We handle the proof state as a tree, with the root node being an equality node. This is one of the theorems in the dataset:
\begin{align*}
    T(T(T(x,T(x,y) \backslash 1),T(x,y) \backslash 1),y) &=\\
    T((T(x,y) \backslash 1) \backslash 1,T(x,y) \backslash 1)
\end{align*}
 The task of the machine learning model is to process the proof state and recognize which actions are most likely to lead to a proof, meaning that the two sides of the starting equation are equal according to the AIMLEAP system.

\section{Methods}
\label{section:training}
This section describes the model training methods, and the next section explains the structure of the neural network models. Here, we briefly explain the approaches that we used as RL baselines, after which we go into detail about the method that we are proposing.

\subsection{RL Baselines}
For comparison, we used implementations of 4 different reinforcement learning baseline methods from the literature. The first method, \textit{Advantage Actor-Critic}, or \textit{A2C} \cite{mnih2016asynchronous} contains ideas on which the other three baseline methods build, so we will go into more detail for this method, while keeping the explanation for the other methods brief. For details, the corresponding papers could be consulted. Reinforcement learning attempts the maximization of the reward obtained when operating in an environment \cite{sutton2018reinforcement}. In our setting, one proof or normalization attempt is an \emph{episode}. Such an episode consists of lists of states $s$ encountered, actions $a$ taken and rewards $r$ received after taking those actions. Our setting is such that there is only a reward of 1 at the end of a successful episode. The lists $a$, $s$ and $r$ are indexed by the timestep $t$.

A2C attempts to find suitable parameters for an agent by minimizing a loss function consisting of two parts:
\begin{align*}
    \mathcal{L} &=  \mathcal{L}_\text{policy}^\text{A2C} +\mathcal{L}_\text{value}^\text{A2C}.
\end{align*}

This loss is calculated by using the states and actions encountered in the environment. The first part of the loss is the policy loss, which for one sample has the form
\begin{align*}
    \mathcal{L}_\text{policy}^\text{A2C} &= - \log \pi_{\theta} (a|s) A(s,a), 
\end{align*}

where $A(s,a)$ is the advantage function and $\pi_{\theta}$ denotes the policy model $\pi$ with parameters $\theta$. $A(s,a)$ denotes the advantage in terms of future rewards of taking action \textit{a} in state \textit{s} compared to the value of the current state. A2C uses the approximation $V_t^n - V(s_t)$ for the advantage, where $V_t^n = \sum_{i=0}^{n-1}\gamma^{i}r_{t+i} + \gamma^{n}V(s_{t+n})$, the $n$-step value estimate. Here $\gamma$ is the discount factor and $n$ is the number of states used for the estimate, indexed by timestep $t$.  The value estimates $V(s)$ for computing the advantage function are supplied by the value predictor $V_{\mu}$ with parameters $\mu$, which is trained using the loss:
\begin{align*}
    \mathcal{L}_\text{value}^\text{A2C} &= \frac{1}{2} (   V_t^n - V_{\mu}(s_t) )^2,
\end{align*}
which minimizes the squared error between the bootstrapped value estimate which takes into account environment rewards and the predicted value. Note that the sets of parameters $\mu$ and $\theta$ may share parameters.

The second RL baseline method we tested in our experiments is ACER, \textit{Actor-Critic with Experience Replay} \cite{wang2016sample}. This approach can make use of data from older episodes to train the current model. ACER applies corrections to the value estimates so that data from old episodes may be used to train the current policy. It also uses trust region policy optimization to limit the size of the policy updates. This method is included as a baseline to check whether making use of a larger replay buffer to update the parameters would be advantageous.

Our third RL baseline is the widely used \textit{proximal policy optimization} (PPO) algorithm \cite{schulman2017proximal}. It restricts the size of the parameter update to avoid causing a large difference between the original model and the updated model. In this way, PPO addresses the training instability of many reinforcement learning approaches. It has been used in various settings, for example complex video games \cite{berner2019dota}. With the potential instability of neural network outputs on tree-structured data, the PPO algorithm is well-positioned. We use the PPO algorithm with clipped objective, as described in \cite{schulman2017proximal}. 

Our final RL baseline uses only the transitions with positive advantage to train on for a portion of the training procedure, to learn more from good episodes. This was proposed as \emph{self-imitation learning} (SIL) \cite{oh2018self}, but to avoid confusion with the method that we are proposing, we expand the acronym to SIL-PAAC, for positive advantage actor-critic.  This algorithm outperformed A2C on the sparse-reward task Montezuma's Revenge. As theorem proving has a sparse reward structure, we included SIL-PAAC as a baseline. More information about the implementations for the baselines can be found in Appendix 4.
\subsection{3SIL}
We introduce stratified shortest solution imitation learning (3SIL) to perform well in the equational theorem proving domain. It learns to explicitly imitate the actions taken during the shortest solutions found for each problem in the dataset. We do this by minimizing the cross-entropy between the model output and the actions taken in the shortest solution. During the procedure no value function is trained. Instead of this, we build upon the assumption for data selection that shorter proofs are better in the context of theorem proving and expression normalization. We keep a history for each problem, where we store the current shortest solution (states seen and actions taken) found for that problem in the training dataset. We can also store multiple shortest solutions for each problem if there are multiple strategies for a proof (the number of solutions kept is governed by the parameter \textit{k}).

During training, we sample state-action pairs from each problem's current shortest solution at an equal probability (if a solution was found). This directly counters one of the phenomena that we had observed: the training examples for the baseline methods tend to be dominated by very long episodes (as they contribute more states and actions). This \textit{stratified} sampling method ensures that problems with short proofs get represented equally in the training process.

The 3SIL algorithm is described in more detail in Algorithm~\ref{alg:example}. Sampling from a noisy version of policy $\pi_{\theta}$ means that actions are sampled from the model-defined distribution and in 5\% of cases a random valid action is selected.

Our approach is similar to the imitation learning algorithm DAGGER (Dataset Aggregation), which was used for several games \cite{ross2011reduction} and modified for branch-and-bound algorithms in \cite{he2014learning}. The behavioral cloning (BC) technique used in robotics \cite{torabi2018behavioral} also shares elements. 3SIL significantly differs from DAGGER and BC because it does not use an outside expert to obtain useful data, because of the stratified sampling procedure and because of the keeping of the shortest solutions for each problem in the training dataset. We include as an additional baseline an implementation of behavioral cloning, where we regard proofs already encountered as coming from an expert. We minimize cross-entropy between the actions in proofs we have found and model predictions. 

For the polynomial and AIM tasks, we introduce two other techniques, biased sampling and episode pruning. In biased sampling, problems without a solution in the history are sampled 5 times more during episode collection than solved problems to accelerate progress. This was determined by testing 1, 2, 5 and 10 as sampling proportions. For episode pruning, when the agent encountered the same state twice, we prune the episode to exclude the looping before storing the episode. This helps the model learn to avoid  these loops.
\begin{algorithm}[tb]
   \caption{CollectEpisode}
   \label{alg:example_func}
\begin{algorithmic}
   \STATE {\bfseries Input:} problem $p$, policy $\pi_{\theta}$,  problem history H
   \STATE Generate episode by following noisy version of $\pi_{\theta}$ on $p$
   \STATE \textbf{If} solution, add list of tuples $(s, a)$ to H[$p$]
   \STATE Keep $k$ shortest solutions in H[$p$]
\end{algorithmic}

\end{algorithm}

\begin{algorithm}[tb]
   \caption{3SIL}
   \label{alg:example}
\begin{algorithmic}
   \STATE {\bfseries Input:} set of problems P, randomly initialized policy $\pi_{\theta}$, batch size $B$, number of batches NB, problem history H, number of warmup episodes $m$, number of episodes $f$, max epochs ME
   \FOR {$e=0$ {\bfseries to} $\text{ME} - 1$}
   \STATE \textbf{if} {$e=0$} \textbf{then} num = $m$ \textbf{else}  num = $f$ \\

   \FOR{$i=0$ {\bfseries to} $\text{num}-1$}
   \STATE CollectEpisode(sample(P), $\pi_{\theta}$, H)  (Algorithm 1)
   
   \ENDFOR
   
   \FOR {$i=0$ {\bfseries to} $\text{NB}-1$ }
   \STATE Sample $B$ tuples ($s$, $a$) with uniform probability for each problem from H
   
   $\text{Update $\pi_{\theta}$ to minimize} - \sum_{b=0}^B \log \pi_{\theta}(a_b | s_b)$
   
   \ENDFOR
   \ENDFOR
\end{algorithmic}

\end{algorithm}

\section{Models}
\label{section:models}

The tree-structured states representing expressions occurring during the tasks will be processed by a neural network. The neural network takes the tree-structured state and predicts an action to take that will bring the expression closer to being normalized or the theorem closer to being proven.

There are two main components to the neural network we use: an \textit{embedding} in the form of a tree neural network that outputs a numerical vector representing the tree-structured proof state and a second \textit{predictor} network that takes this vector representation of the state and outputs a distribution of the actions possible in the environment (in the reinforcement learning baselines that we use, this predictor network has the additional task of predicting the value of a state).

Tree neural networks have been used in natural language processing \cite{irsoy2014deep} and also in Robinson arithmetic expression embedding \cite{gauthier2020tree}. These networks consist of smaller neural networks, each representing one of the possible functions that occur in the expressions. For example, there will be separate networks representing addition and multiplication. The cursor is a special unary operation node with its own network that we insert into the tree at the current location. For each unique leaf, such as the constant 0, the variables \textit{x, y, z}, or identity element 1,  we generate a random vector (from a standard normal distribution) that will represent this leaf. Aside from the 0 in the RA experiments, these vectors are trainable.

The numerical representation of a tree is constructed by starting at the leaves of the tree, for which we can look up the generated vectors. These vectors act as input to the neural networks that represent the parent node's operation, yielding a new vector, which now represents the subtree of the parent node. The process repeats until there is a single vector for the entire tree after the root node is processed.

The neural networks representing each operation consist of a linear transformation, a non-linearity in the form of a rectified linear unit (ReLU) and another linear transformation. In the case of binary operations, the first linear transformation will have an input dimension of \textit{2n} and an output dimension of \textit{n}, where \textit{n} is the dimension of the vectors representing the leaves of the tree (the \textit{internal representation size}).

When we have obtained a single vector embedding representing the entire tree data structure, this vector serves as the input to the \textit{predictor} neural network, which consists of three linear layers, with non-linearities (Sigmoid/ReLU) in between these layers. The last layer has an output dimension equal to the number of possible actions in the environment. We obtain a probability distribution over the actions by applying the softmax function to the output of this last layer. In the cases where we also need a value prediction, there is a parallel last layer that predicts the state's value (usually referred to as a \textit{two-headed} network \cite{silver2017mastering}).

The internal representation size $n$ for the Robinson arithmetic experiments is set to 16, for the polynomial and AIM tasks this is 32. The number of neurons in each layer (except for the last one) of the predictor networks is 64. 

In the AIM dataset task, an arbitrary number of variables can be introduced during the proof. These are represented by untrainable random vectors. We add a special neural network (with the same architecture as the networks representing unary operations, so from size \textit{n} to \textit{n}) that processes these vectors before 
they are processed by the rest of the tree neural network embedding. The idea is that this neural network learns to project these new variable vectors into a subspace and that an arbitrary number of variables can be handled. The vectors are resampled at the start of each episode, so the agent cannot learn to recognize variables. This approach was inspired by the \emph{prime} mechanism in \cite{gauthier2020tree}, but we use separate vectors for all  variables instead of building vectors sequentially.

\section{Arithmetic Experiments}
\label{section:ar_exp}
\subsection{Robinson Normalization}
The Robinson arithmetic dataset \cite{gauthier2019deep} is split into three distinct sets, based on the number of steps that it takes a fixed rewriting strategy to normalize the expression. This fixed strategy, LOPL, which stands for \textit{left outermost proof length}, always rewrites the leftmost possible element. If it takes this strategy less than 90 steps to solve the problem, it is in the \textit{low} difficulty category. Problems with a difficulty between 90 and 130 are in the \textit{medium} category and a greater difficulty than 130 leads to the \textit{high} category. The \textit{high} dataset also contains problems the LOPL strategy could not solve within the timelimit. The \textit{low} dataset is split into a training and testing set. We train on the \textit{low} difficulty problems, but after training we also test on problems with a higher difficulty. Because we have a difficulty measure for this dataset, we use a curriculum setup. We start by learning to normalize the expressions that a fixed strategy can normalize in the smallest number of steps. This setup is similar to \cite{gauthier2019deep}.

The 400 problems with the lowest difficulty are the starting point. Every time an agent reaches 95 percent success rate when evaluated on a sample of size 400 from these problems, we add 400 more difficult problems to set of  training problems $P$. Agents are evaluated after every epoch. The blocks of size 400 are called \emph{levels}. The number of episodes $m$ and $f$ are set to 1000. For 3SIL and BC, the batch size $\text{BS}$ is 32 and the number of batches $\text{NB}$ is 250. The baselines are configured so that the number of episodes and training transitions is at least as many as the 3SIL/BC approaches. For more details on this, see Appendix 4. If an episode takes more than 100 steps, it is stopped. ADAM is used as an optimizer. For 3SIL and BC, the evaluation is done greedily (always take the highest probability actions). For the other methods, we performed experiments with both greedy and non-greedy (sample from the model distribution and add 5\% noise) evaluation and show the results of the configuration which performed best (which in most cases was the non-greedy evaluation, except for PPO).
\begin{figure}[t]
\vskip 0.1in
\begin{center}
\centerline{\includegraphics[width=0.6\columnwidth]{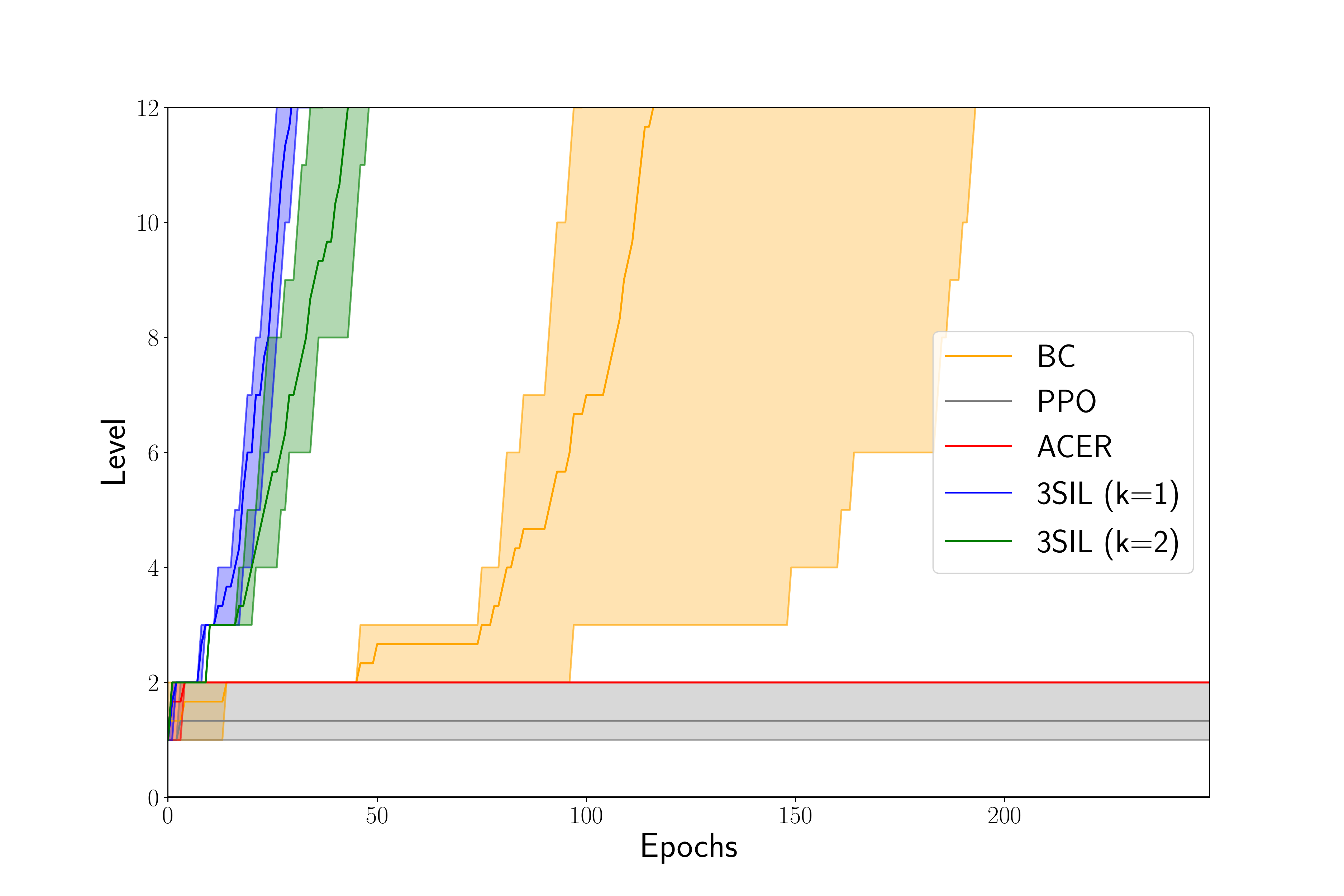}}
\caption{The level in the curriculum reached by each method. Each method was run three times. The bold line shows the mean performance and the shaded region shows the minimum and maximum performance.}
\label{fig:training_with_baseline}
\end{center}

\end{figure}

In Figure \ref{fig:training_with_baseline}, we show the progression through the training curriculum for behavioral cloning, the RL methods and two configurations of 3SIL. Behavioral cloning simply imitates actions from successful episodes. Of the RL baselines, PPO reaches the second level in one run, while ACER steadily solves the first level and in the best run solves around 80\% of the second level. Both methods do not learn enough solutions for the second level to advance to the third. A2C and SIL-PAAC do not reach the second level, so these are left out of the plot. However, they do learn to solve about 70-80\% of the the first 400 problems. From these results we can conclude that the RL methods do not perform well on this task in our experiment.  We attribute this to the difficulty of learning a stable value function with the tree neural network as an embedding and the sparse rewards. Our hypothesis is that because this value estimate influences the policy updates, the RL methods do not learn well on this task with the current setup. Note that the two methods with a trust region update mechanism, ACER and PPO, perform better than the methods without this mechanism. From these results, it is clear that 3SIL with $k=1$ is the best-performing configuration. It reaches the end of the training curriculum of about 5000 problems in 40 epochs. 

While our approach works well on the training set, we must check if the models generalize to unseen examples. Only the methods that reached the end of the curriculum are tested. In Table \ref{tab:robinson_test_set}, we show the results of evaluating the performance of our models on the three different test sets. Because we expect to require longer solutions, the episode limits are expanded from 100 steps to 200 and 250 for the \textit{medium} and \textit{high} datasets respectively. For the \textit{low} and \textit{medium} datasets, the second of which contains problems with more difficult solutions than the training data, the models solve almost all test problems. For the \textit{high} difficulty dataset, the performance drops by at least 20 percentage points. Our method outperforms the Monte Carlo Tree Search approach used in \cite{gauthier2019deep} on the same datasets.

\begin{table}[tb]
\caption{Generalization performance with greedy evaluation for test set problems for the Robinson arithmetic normalization tasks. Generalization is high on the low and medium difficulty datasets (the agents are trained on problems of the same difficulty as the low difficulty dataset). For the high difficulty dataset, performance drops.}
\label{tab:robinson_test_set}
\vskip 0.15in
\begin{center}
\begin{small}
\begin{sc}
\begin{tabular}{llll}
\toprule
{} &          Low &       Medium &         High \\
\midrule
3SIL (k=1) &  1.00 $\pm$ 0.01 &  0.98 $\pm$ 0.03 &  0.77 $\pm$ 0.10 \\
3SIL (k=2) &  0.99 $\pm$ 0.00 &  0.96 $\pm$ 0.01 &  0.66 $\pm$ 0.08 \\
BC &  0.98 $\pm$ 0.01 &  0.98 $\pm$ 0.01 &  0.56 $\pm$ 0.05 \\
\bottomrule
\end{tabular}
\end{sc}
\end{small}
\end{center}

\end{table}

\subsection{Polynomial Normalization}
We also tested our method on the more challenging polynomial normalization task. We introduce two extra techniques, biased problem sampling and episode pruning, that improve the results (see Section 4.2). These techniques are used on top of the best performing configuration on the Robinson arithmetic dataset (3SIL with $k=1$). We now train for 750 epochs and the next level is reached when 90\% of the evaluation problems are solved.

We use an even mixture of the six datasets provided in \cite{piotrowski2019can}, with 1000 samples from each dataset. Again, we measure the difficulty of each problem as the number of steps that a fixed strategy takes. For the polynomial normalization, we counted the number of steps needed by the \textit{expand} function of the mathematical Python library \textit{Sympy} \cite{10.7717/peerj-cs.103}. For testing, we take 1000 problems from each of the six test sets.

\begin{figure}[b!]
\vskip 0.1in
\begin{center}
\centerline{\includegraphics[trim=0cm 0cm 0cm 3cm,clip=true, width=0.6\columnwidth]{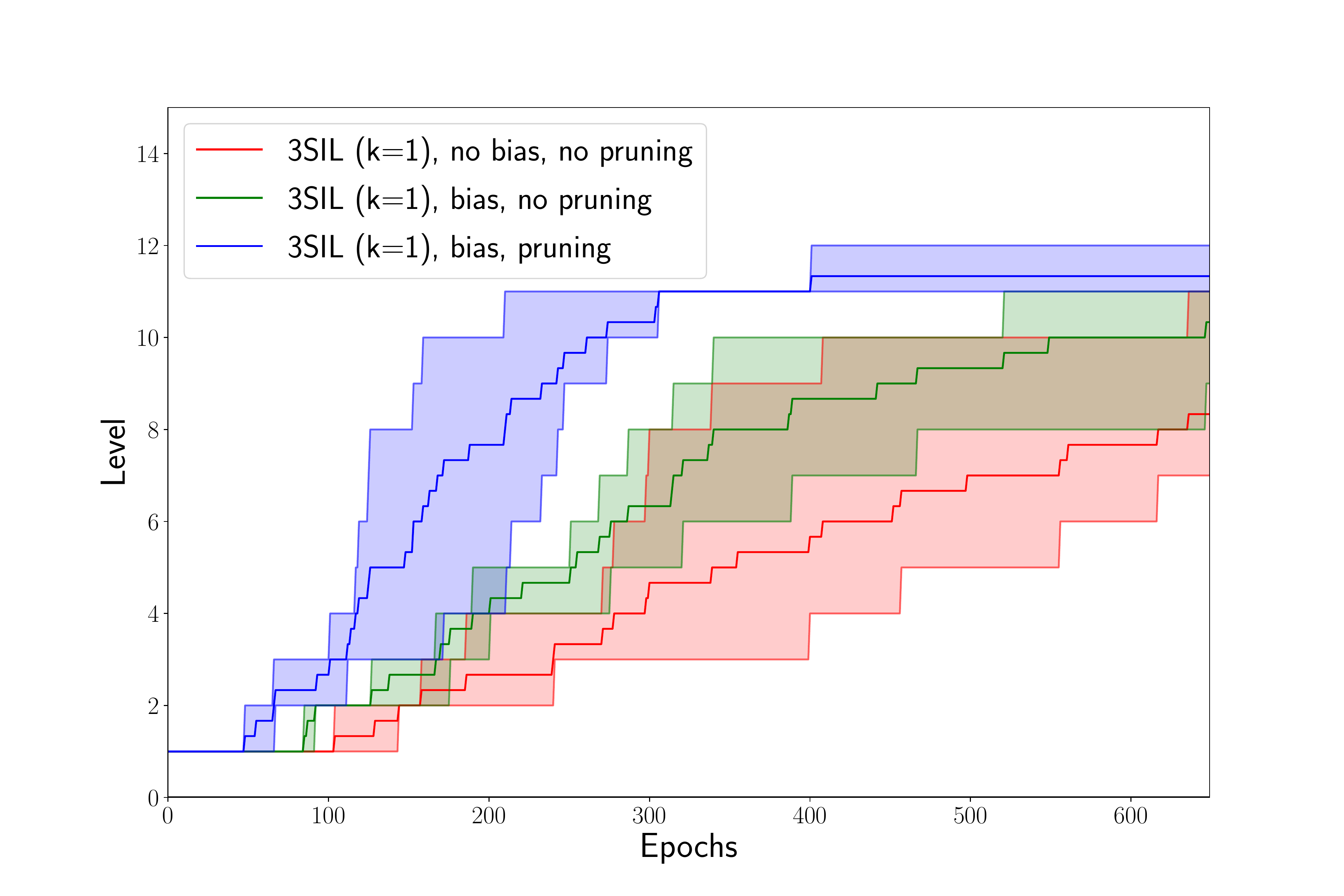}}
\caption{3SIL algorithm training progress on the polynomial normalization task. Both the unsolved problem sampling bias and the loop pruning of episodes improve learning on the polynomial curriculum. }
\label{fig:training_polynomial}
\end{center}
\vskip -0.2in
\end{figure}

In Figure \ref{fig:training_polynomial}, we show the training progress for three different configurations. The results show that both bias and pruning leads to a faster progression on the curriculum.
The generalization performance of the method is shown in Table \ref{tab:polynomial_test_set}. The success rate on the polynomial dataset is worse than on the Robinson dataset, but this is expected as the task is more difficult. We are solving a much more challenging task than \cite{piotrowski2019can} with the same dataset, as we need to supply a proof instead of doing a one-shot transformation, but still can solve 63\% of test problems. Especially the successor representations of large numbers make the task hard for our model. The exponentiation operation can cause large numbers to appear, which are represented as deep trees.

\begin{table}[tb]
\caption{Test set performance of 3SIL (k=1) as fraction of problems solved for the polynomial rewriting task with the biased sampling and loop pruning extensions. The modifications lead to better test set performance. }
\label{tab:polynomial_test_set}
\vskip 0.15in
\begin{center}
\begin{small}
\begin{sc}
    \begin{tabular}{ll}
\toprule
{} &        Test Set Performance \\
\midrule
 No Bias, No Pruning     &  0.56 $\pm$ 0.04 \\
Bias, No Pruning      &  0.58 $\pm$ 0.03 \\
Bias, Pruning       &  0.63 $\pm$ 0.02 \\
\bottomrule
\end{tabular}
\end{sc}
\end{small}
\end{center}
\vskip -0.1in
\end{table}

\section{AIM Dataset Experiments}
\label{section:aim_exp}
We will now show our results on the AIM Conjecture dataset. The setup is similar to the one for the polynomial expression normalization tasks. We apply 3SIL ($k=1$) to train models in the AIMLEAP environment. Ten percent of the AIM dataset is used as a hold-out test set, not seen during training. As there is no estimate for the difficulty of the problems in terms of the actions available to the model, we do not use a curriculum ordering for these experiments. The number of episodes collected before training $m$ is 2,000,000. These random proof attempts result in about 300 proofs. The model learns from these proofs and afterwards the search for new proofs is also guided by the model's predictions. For the AIM experiments, episodes are stopped after 30 steps in the AIMLEAP environment. The models are trained for 100 epochs. The number of collected episodes per epoch $f$ is 10,000. The successful proofs are stored, and the shortest proof for each theorem is kept. $\text{NB}$ is 500 and $\text{BS}$ is set to 32. The number of problems with a solution in the history after each epoch of the training run is shown in Appendix 6.

After 100 epochs, about 2500 of around 3100 problems in the training dataset have a solution in their history. To test the generalization capability of the models, we inspect their performance on the holdout test set problems. In Table \ref{tab:test_success_rate}  we compare the success rate of the trained models on a separate test set with three different automated theorem provers: E, Waldmeister and Prover9. E is currently one of the best overall automated theorem provers \cite{Sut20-CASC}, Waldmeister is a prover specialized in memory-efficient equational theorem proving \cite{Hillenbrand2004WALDMEISTERH} and Prover9 is the theorem prover that is used for AIM conjecture research and the prover that the dataset was generated by. Waldmeister and E are the best performing solvers in competitions for the relevant unit equality (UEQ) category \cite{Sut20-CASC}. 

\begin{figure}[t!]
\vskip 0.2in
\begin{center}
\centerline{\includegraphics[trim=2cm 4.5cm 2cm 2.5cm,clip=true, width=0.6\columnwidth]{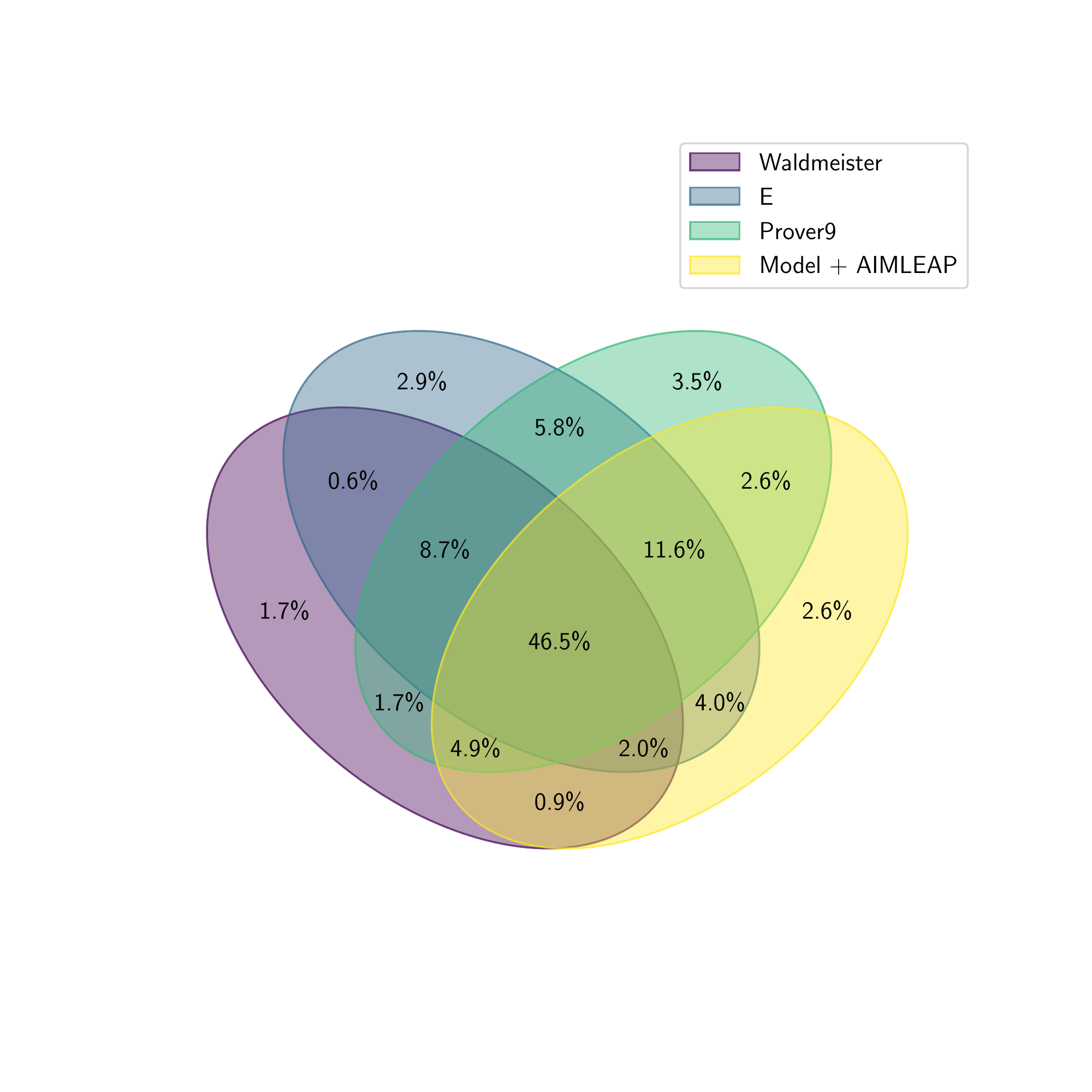}}
\caption{Venn diagram of the test set problems solved by each solver with 60s time limit.}
\label{fig:venn}
\end{center}
\vskip -0.2in
\end{figure}

The results show that a single greedy evaluation of the model is not as strong as the theorem proving software. However, the theorem provers got 60 seconds of execution time, and the execution of the model, including interaction with AIMLEAP, takes on average less than 1 second. We allowed the model to use 60 seconds, by running attempts until the time was up, sampling actions from the model distribution with 5\% noise, instead of using greedy execution. With this approach, the model outperforms Waldmeister. 

Figure \ref{fig:venn} shows the overlap between the problems solved by each prover. The diagram shows that each theorem prover found a few solutions that no other prover could find within the time limit. Almost half of all problems from the test set that are solved are solved by all four systems.

\begin{table}[b!]
\caption{Theorem proving performance on the hold-out test set in fraction of problems solved. Means and standard deviations are the results of evaluations of 3  different models from 3 different training runs.}
\label{tab:test_success_rate}
\vskip 0.15in
\begin{center}
\begin{small}
\begin{sc}
   \begin{tabular}{lr}
\toprule
     Method &  Success Rate \\
\midrule
           E (60s) &          0.802 \\
            Waldmeister (60s) &          0.655 \\
     Prover9 (60s)  &         0.833 \\
       Model (1x) &           0.586 $\pm$ 0.029 \\
       Model (60s) &         0.702 $\pm$ 0.015 \\
\bottomrule
\end{tabular}
\end{sc}
\end{small}
\end{center}
\vskip -0.1in
\end{table}
 We also combine the model with \textit{Prover9}. The model modifies the starting form of the goal, for a maximum of 1 second in the AIMLEAP environment. This produces new expression on one or both sides of the equality. We then add, as lemmas, equalities between the left-hand side of the goal before the model's rewriting and after the rewriting. The same is done for the right-hand side. For each problem, this procedure yields new lemmas that are added to the problem specification file that is given to \textit{Prover9}.
 \begin{table}[tb]
 
    \caption{Prover 9 theorem proving performance on the hold-out test set when injecting lemmas suggested by the learned model. \textit{Prover9}'s performance increases when using the model's suggested lemmas.}
    \label{tab:injection_success_rate}
\vskip 0.15in
\begin{center}
\begin{small}
\begin{sc}
   \begin{tabular}{lr}
\toprule
     Method &  Success Rate \\
\midrule
        Prover9 (1s) &          0.715 \\
        Prover9 (2s) &          0.746 \\
        Prover9 (1s) + Model Lemmas (1s) &          0.841 $\pm$ 0.019 \\
\bottomrule
\end{tabular}
\end{sc}
\end{small}
\end{center}
\vskip -0.1in
\end{table}

In Table \ref{tab:injection_success_rate}, it is shown that adding lemmas suggested by the rewriting actions of the trained models improves the performance of \textit{Prover9}. Running \textit{Prover9} for 2 seconds results in better performance than running it for 1 second, as expected. The combined system improved on \textit{Prover9's} 2-second performance, indicating that the model suggests useful lemmas. The best configuration even outperforms Prover9 with a 60 second time limit (Table \ref{tab:test_success_rate}). These results indicate that using a learned model in combination with a  standard theorem prover is possible and a viable direction.
 \section{Discussion}
\label{section:disc}
The experiments on the Robinson arithmetic task indicate that 3SIL outperforms the standard RL algorithms we tested. This could be due to the difficulty of learning a stable value function in an environment where the rewards are sparse and the input data has a tree structure. In some cases, a destabilizing update  decreased performance. Our method, without a value function, exhibited more stable behavior.

Generalization performance for the tasks is high. We conclude that the structure and relatively low number of parameters in the neural networks prevent overfitting. We experimented with the transformer architectures to process the tree-structured input, as in \cite{lample2019deep}, but the training of this model was unsuccessful. This is probably due to the low amount of data available at the start of training being insufficient for the transformer.
\section{Conclusion and Future Work}
\label{section:conc}
Our experiments show that it is possible to use 3SIL to learn equational theorem proving on theorems related to the \textit{Abelian Inner Mapping Conjecture}. The model outperforms some established theorem provers, indicating that it is possible for learned provers to compete with standard theorem provers on specific domains or serve as an empowering tool for existing provers. The general case of theorem proving is undecidable, indicating that the models are solving hard problems.

In future work, we will apply our method to equational reasoning tasks for other group-theory related mathematical structures. In addition, we plan to investigate the extent to which \textit{Prover9} can be assisted by our method. An especially interesting research direction concerns selecting which proofs to learn from: some proofs might use sub-proofs that are more general than other sub-proofs. 

\section{Acknowledgements}
We would like to thank Chad E. Brown for helping us use AIMLEAP as a reinforcement learning environment and for his help with running the automated theorem provers. We would also like to thank Thibault Gauthier and Bartosz Piotrowski for useful comments. 

The results were supported by the Ministry of Education, Youth and Sports within the dedicated program ERC CZ under the project POSTMAN with reference LL1902, by the RICAIP project that has received funding from the European Union's Horizon 2020 research and innovation programme under grant agreement No 857306, by the
    ERC Consolidator grant AI4REASON
    no.~649043, by the Czech project AI\&Reasoning
    CZ.02.1.01/0.0/0.0/15\_003/0000466 and by the European Regional
    Development Fund.

\clearpage

\printbibliography

@inproceedings{he2016deep,
  title={Deep residual learning for image recognition},
  author={He, Kaiming and Zhang, Xiangyu and Ren, Shaoqing and Sun, Jian},
  booktitle={Proceedings of the IEEE Conference on Computer Vision and Pattern Recognition},
  pages={770--778},
  year={2016}
}

@article{mnih2015human,
  title={Human-level control through deep reinforcement learning},
  author={Mnih, Volodymyr and Kavukcuoglu, Koray and Silver, David and Rusu, Andrei A and Veness, Joel and Bellemare, Marc G and Graves, Alex and Riedmiller, Martin and Fidjeland, Andreas K and Ostrovski, Georg and others},
  journal={Nature},
  volume={518},
  number={7540},
  pages={529--533},
  year={2015},
  publisher={Nature Publishing Group}
}

@inproceedings{graves2006connectionist,
  title={Connectionist temporal classification: labelling unsegmented sequence data with recurrent neural networks},
  author={Graves, Alex and Fern{\'a}ndez, Santiago and Gomez, Faustino and Schmidhuber, J{\"u}rgen},
  booktitle={Proceedings of the 23rd International Conference on Machine Learning},
  pages={369--376},
  year={2006}
}

@inproceedings{chvalovsky2019enigma,
  title={ENIGMA-NG: efficient neural and gradient-boosted inference guidance for E},
  author={Chvalovsk{\'y}, Karel and Jakubuv, Jan and Suda, Martin and Urban, Josef},
  booktitle={International Conference on Automated Deduction},
  pages={197--215},
  year={2019},
  organization={Springer}
}

@article{kaliszyk2018reinforcement,
  title={Reinforcement learning of theorem proving},
  author={Kaliszyk, Cezary and Urban, Josef and Michalewski, Henryk and Ol{\v{s}}{\'a}k, Miroslav},
  journal={Advances in Neural Information Processing Systems},
  volume={31},
  pages={8822--8833},
  year={2018}
}

@inproceedings{bansal2019holist,
  title={{HOList}: An environment for machine learning of higher order logic theorem proving},
  author={Bansal, Kshitij and Loos, Sarah and Rabe, Markus and Szegedy, Christian and Wilcox, Stewart},
  booktitle={International Conference on Machine Learning},
  pages={454--463},
  year={2019}
}

@article{gauthier2020tactictoe,
  title={{TacticToe}: Learning to Prove with Tactics},
  author={Gauthier, Thibault and Kaliszyk, Cezary and Urban, Josef and Kumar, Ramana and Norrish, Michael},
  journal={Journal of Automated Reasoning},
  pages={1--30},
  year={2020},
  publisher={Springer}
}

@article{phillips2010automated,
  title={Automated theorem proving in quasigroup and loop theory},
  author={Phillips, JD and Stanovsk{\'y}, David},
  journal={Ai Communications},
  volume={23},
  number={2-3},
  pages={267--283},
  year={2010},
  publisher={IOS Press}
}

@inproceedings{de2008z3,
  title={Z3: An efficient {SMT} solver},
  author={De Moura, Leonardo and Bj{\o}rner, Nikolaj},
  booktitle={International conference on Tools and Algorithms for the Construction and Analysis of Systems},
  pages={337--340},
  year={2008},
  organization={Springer}
}

@incollection{kinyon2013loops,
  title={Loops with abelian inner mapping groups: An application of automated deduction},
  author={Kinyon, Michael and Veroff, Robert and Vojt{\v{e}}chovsk{\'y}, Petr},
  booktitle={Automated Reasoning and Mathematics},
  pages={151--164},
  year={2013},
  publisher={Springer}
}

@article{gauthier2019deep,
  title={Deep reinforcement learning in {HOL4}},
  author={Gauthier, Thibault},
  journal={arXiv preprint arXiv:1910.11797},
  url={https://arxiv.org/abs/1910.11797v1},
  year={2019}
}

@article{burris1993tarski,
  title={Tarski's high school identities},
  author={Burris, Stanley and Lee, Simon},
  journal={The American Mathematical Monthly},
  volume={100},
  number={3},
  pages={231--236},
  year={1993},
  publisher={Taylor \& Francis}
}

@article{piotrowski2019can,
  title={Can Neural Networks Learn Symbolic Rewriting?},
  author={Piotrowski, Bartosz and Urban, Josef and Brown, Chad E and Kaliszyk, Cezary},
  journal={ICML Workshop on Learning and Reasoning with Graph-Structured Data},
  year={2019}
}

@article{brown2020learning,
  title={Learning to Advise an Equational Prover},
  author={Brown, Chad E and Piotrowski, Bartosz and Urban, Josef},
  year={2020},
  journal={Artificial Intelligence and Theorem Proving}
}

@inproceedings{mnih2016asynchronous,
  title={Asynchronous methods for deep reinforcement learning},
  author={Mnih, Volodymyr and Badia, Adria Puigdomenech and Mirza, Mehdi and Graves, Alex and Lillicrap, Timothy and Harley, Tim and Silver, David and Kavukcuoglu, Koray},
  booktitle={International conference on machine learning},
  pages={1928--1937},
  year={2016}
}

@article{wang2016sample,
  title={Sample efficient actor-critic with experience replay},
  author={Wang, Ziyu and Bapst, Victor and Heess, Nicolas and Mnih, Volodymyr and Munos, Remi and Kavukcuoglu, Koray and de Freitas, Nando},
  journal={International Conference on Learning Representations},
  year={2016}
}

@article{schulman2017proximal,
  title={Proximal policy optimization algorithms},
  author={Schulman, John and Wolski, Filip and Dhariwal, Prafulla and Radford, Alec and Klimov, Oleg},
  journal={arXiv preprint arXiv:1707.06347},
  year={2017}
}

@inproceedings{oh2018self,
  title={Self-Imitation Learning},
  author={Oh, Junhyuk and Guo, Yijie and Singh, Satinder and Lee, Honglak},
  booktitle={International Conference on Machine Learning},
  pages={3878--3887},
  year={2018}
}

@article{berner2019dota,
  title={Dota 2 with large scale deep reinforcement learning},
  author={Berner, Christopher and Brockman, Greg and Chan, Brooke and Cheung, Vicki and Debiak, Przemys{\l}aw and Dennison, Christy and Farhi, David and Fischer, Quirin and Hashme, Shariq and Hesse, Chris and others},
  journal={arXiv preprint arXiv:1912.06680},
  year={2019}
}

@inproceedings{lample2019deep,
  title={Deep Learning For Symbolic Mathematics},
  author={Lample, Guillaume and Charton, Fran{\c{c}}ois},
  booktitle={International Conference on Learning Representations},
  year={2019}
}

@article{silver2016mastering,
  title={Mastering the game of Go with deep neural networks and tree search},
  author={Silver, David and Huang, Aja and Maddison, Chris J and Guez, Arthur and Sifre, Laurent and Van Den Driessche, George and Schrittwieser, Julian and Antonoglou, Ioannis and Panneershelvam, Veda and Lanctot, Marc and others},
  journal={Nature},
  volume={529},
  number={7587},
  pages={484--489},
  year={2016},
  publisher={Nature Publishing Group}
}

@inproceedings{gauthier2020deep,
  title={Deep Reinforcement Learning for Synthesizing Functions in Higher-Order Logic.},
  booktitle={International Conference on Logic for Programming, Artificial Intelligence and Reasoning},
  author={Gauthier, Thibault},
  year={2020}
}

@inproceedings{torabi2018behavioral,
author = {Torabi, Faraz and Warnell, Garrett and Stone, Peter},
title = {Behavioral Cloning from Observation},
year = {2018},
isbn = {9780999241127},
publisher = {AAAI Press},
booktitle = {Proceedings of the 27th International Joint Conference on Artificial Intelligence},
pages = {4950–4957},
numpages = {8},
location = {Stockholm, Sweden},
series = {IJCAI'18}
}

@inproceedings{ross2011reduction,
  title={A reduction of imitation learning and structured prediction to no-regret online learning},
  author={Ross, St{\'e}phane and Gordon, Geoffrey and Bagnell, Drew},
  booktitle={Proceedings of the 14th International Conference on Artificial Intelligence and Statistics},
  pages={627--635},
  year={2011}
}

@article{he2014learning,
  title={Learning to search in branch and bound algorithms},
  author={He, He and Daume III, Hal and Eisner, Jason M},
  journal={Advances in Neural Information Processing Systems},
  volume={27},
  pages={3293--3301},
  year={2014}
}

@article{irsoy2014deep,
  title={Deep recursive neural networks for compositionality in language},
  author={Irsoy, Ozan and Cardie, Claire},
  journal={Advances in Neural Information Processing Systems},
  volume={27},
  pages={2096--2104},
  year={2014}
}

@inproceedings{gauthier2020tree,
  title={Tree Neural Networks in {HOL4}},
  author={Gauthier, Thibault},
  booktitle={International Conference on Intelligent Computer Mathematics},
  pages={278--283},
  year={2020},
  organization={Springer}
}

@article{10.7717/peerj-cs.103,
 title = {SymPy: symbolic computing in {Python}},
 author = {Meurer, Aaron and Smith, Christopher P. and Paprocki, Mateusz and \v{C}ert\'{i}k, Ond\v{r}ej and Kirpichev, Sergey B. and Rocklin, Matthew and Kumar, AMiT and Ivanov, Sergiu and Moore, Jason K. and Singh, Sartaj and Rathnayake, Thilina and Vig, Sean and Granger, Brian E. and Muller, Richard P. and Bonazzi, Francesco and Gupta, Harsh and Vats, Shivam and Johansson, Fredrik and Pedregosa, Fabian and Curry, Matthew J. and Terrel, Andy R. and Rou\v{c}ka, \v{S}t\v{e}p\'{a}n and Saboo, Ashutosh and Fernando, Isuru and Kulal, Sumith and Cimrman, Robert and Scopatz, Anthony},
 year = 2017,
 month = jan,
 volume = 3,
 pages = {e103},
 journal = {PeerJ Computer Science},
 issn = {2376-5992},
 url = {https://doi.org/10.7717/peerj-cs.103},
 doi = {10.7717/peerj-cs.103}
}

@book{sutton2018reinforcement,
  title={Reinforcement learning: An introduction},
  author={Sutton, Richard S and Barto, Andrew G},
  year={2018}
}

@article{silver2017mastering,
  title={Mastering the game of go without human knowledge},
  author={Silver, David and Schrittwieser, Julian and Simonyan, Karen and Antonoglou, Ioannis and Huang, Aja and Guez, Arthur and Hubert, Thomas and Baker, Lucas and Lai, Matthew and Bolton, Adrian and others},
  journal={Nature},
  volume={550},
  number={7676},
  pages={354--359},
  year={2017},
  publisher={Nature Publishing Group}
}

@Article{Sut20-CASC,
    Author       = "Sutcliffe, G.",
    Year         = "2020",
    Title        = {The {CADE-27} Automated Theorem Proving System Competition 
                    - {CASC-27}},
    Journal      = "AI Communications",
    Volume       = "32",
    Number       = "5-6",
    Pages        = "373-389"
}

@article{Hillenbrand2004WALDMEISTERH,
  title={WALDMEISTER - High-Performance Equational Deduction},
  author={T. Hillenbrand and A. Buch and R. Vogt and Bernd L{\"o}chner},
  journal={Journal of Automated Reasoning},
  year={2004},
  volume={18},
  pages={265-270}
}

@unpublished{prover9-mace4,
    author = {W. McCune},
    title = {Prover9 and {Mace}},
    url = {http://www.cs.unm.edu/~mccune/prover9/},
    year = {2010}}

@article{DBLP:journals/jar/Veroff96,
  author    = {Robert Veroff},
  title     = {Using Hints to Increase the Effectiveness of an Automated Reasoning
               Program: Case Studies},
  journal   = {J. Autom. Reason.},
  volume    = {16},
  number    = {3},
  pages     = {223--239},
  year      = {1996},
  url       = {https://doi.org/10.1007/BF00252178},
  doi       = {10.1007/BF00252178},
  bibsource = {dblp computer science bibliography, https://dblp.org}
}
\clearpage
\section*{Appendix}
\setcounter{section}{0}


\section{Robinson arithmetic actions}
Agents in the Robinson arithmetic environment can use these actions to modify the expressions. After application of actions 1-7, the cursor is reset to the root of the expression.
\begin{multicols}{2}
\begin{enumerate}
    \item $x + 0 \to x $
    \item $x \to x + 0 $
    \item $x + S(y) \to S(x + y) $
    \item $S(x + y) \to x + S(y)$
    \item $x * 0 \to 0$
    \item $x * S(y) \to x * y + x$
    \item $x * y + x \to x * S(y)$
    \item Move cursor to left child
    \item Move cursor to right child
\end{enumerate}
\end{multicols}

\section{Polynomial arithmetic actions}
Agents in the Polynomial arithmetic environment can use these actions to modify the expressions. After any application that is not action 8 or 9, the cursor is reset to the root of the expression.
\begin{multicols}{2}
\begin{enumerate}
    \item $x + 0 \to x $
    \item $x \to x + 0 $
    \item $x + S(y) \to S(x + y) $
    \item $S(x + y) \to x + S(y)$
    \item $x * 0 \to 0$
    \item $x * S(y) \to x * y + x$
    \item $x * y + x \to x * S(y)$
    \item Move cursor to left child
    \item Move cursor to right child
    \item $x + y \to y + x$ or $x * y \to y * x$
    \item $x^0 \to 1$
    \item $x^{S(y)} \to x ^ y * x$
    \item $x ^ y * x \to x ^{S(y)}$
    \item $(x + y) + z \to x + (y + z)$ or  $(x * y) * z \to x * (y * z)$
    \item $x + (y + z) \to (x + y) + z$ or $x * (y * z) \to (x * y) * z$
    \item $x * (y + z) \to x * y + x * z$
    \item $x * y + x * z \to x * (y + z)$
    \item $x * 1 \to x$
    \item $x \to x * 1$
    \item $1 ^ x \to 1$
    \item $x ^ 1 \to x$
    \item $x \to x ^ 1$
    \item $x ^ {(y + z)} \to x ^ y * x ^ z$
    \item $x ^ y * x ^ z \to x ^ {(y + z)}$
    \item $(x * y) ^ z \to x ^ z * y ^ z$
    \item $x ^ z * y ^ z \to (x * y) ^ z$
    \item $(x ^ y) ^ z \to x ^ {(y * z)}$
    \item $ x ^ {(y * z)}  \to (x ^ y) ^ z$
    
\end{enumerate}
\end{multicols}
\section{AIM Dataset actions}
In the AIMLEAP environment, the agents are able to use the following axioms, definitions, identities and known propositions to modify the current expression. Note that the agent may apply these in either direction if applicable. In addition, the agent may move the cursor to the first, second or third argument of the current node.
\begin{multicols}{2}

Axiom lid : $e * x = x. $ \\

Axiom rid : $x * e = x. $ \\

Axiom b1 : $x \setminus (x * y) = y.$ \\

Axiom b2 : $x * (x \setminus y) = y. $ \\

Axiom s1 : $(x * y) / y = x. $ \\

Axiom s2 : $(x / y) * y = x. $ \\

Definition $a(x,y,z) := (x*(y*z)) \setminus ((x*y)*z).$ \\

Definition $K(x,y) := (y*x)\setminus(x*y).$ \\

Definition $T(u,x) := x \setminus (u*x).$\\

Definition $L(u,x,y) := (y*x) \setminus (y*(x*u)).$\\

Definition $R(u,x,y) := ((u*x)*y)/(x*y). $\\

Axiom TT: $T(T(u,x),y) = T(T(u,y),x).$\\

Axiom TL: $T(L(u,x,y),z) = L(T(u,z),x,y).$\\

Axiom TR: $T(R(u,x,y),z) = R(T(u,z),x,y).$\\

Axiom LR: $L(R(u,x,y),z,w) = R(L(u,z,w),x,y)$\\

Axiom LL: $L(L(u,x,y),z,w) = L(L(u,z,w),x,y).$\\

Axiom RR: $R(R(u,x,y),z,w) = R(R(u,z,w),x,y).$\\

Known id1: $y / (x \setminus y) = x.$\\

Known id2: $(y / x) \setminus y = x.$\\

Known id3: $e \setminus x = x.$\\

Known id4: $x / e = x.$\\

Known id5: $x \setminus x = e.$\\

Known id6: $x / x = e.$\\

Known prop\_034cf5c5: $x * T(y,x) = y * x.$\\

Known prop\_66f8dd43: $T(x / y,y) = y \setminus x.$\\

Known prop\_81894ca4: $(x * T(y,x)) / x = y.$\\

Known prop\_da4738e2: $T(x,x \setminus y) = (x \setminus y) \setminus y.$\\

Known prop\_4a90a23f: $x * T(T(y,x),z) = T(y,z) * x.$\\

Known prop\_73fa4877: $T(T(x / y,z),y) = T(y \setminus x,z).$\\

Known prop\_d10a3b1a: $(x * y) * L(z,y,x) = x * (y * z).$\\

Known prop\_293cbc16: $L(x \setminus y,x,z) = (z * x) \setminus (z * y).$\\

Known prop\_55464ec9: $R(x,y,z) * (y * z) = (x * y) * z.$\\

Known prop\_61fb8127: $R(x / y,y,z) = (x * z) / (y * z).$\\

Known prop\_ddd1c86f: $x * ((x \setminus e) * y) = L(y,x \setminus e,x).$\\

Known prop\_0d7e7151: $(x \setminus e) * y = x \setminus L(y,x \setminus e,x).$\\

Known prop\_1aae4a83: $x * L(T(y,x),z,w) = L(y,z,w) * x.$\\

Known prop\_1db0183a: $x * R(T(y,x),z,w) = R(y,z,w) * x.$\\

Known prop\_a4abb1e0: $T(R(x / y,z,w),y) = R(y \setminus x,z,w).$\\

Known prop\_55842885: $T(x / y,z) * y = y * T(y \setminus x,z).$\\

Known prop\_1a725917: $(x * y) \setminus (x * (z * y)) = L(T(z,y),y,x).$\\

Known prop\_c626af2d: $T(x,x * y) = L(T(x,y),y,x).$\\

Known prop\_526a359c: $T((x * y) / (z * y),z) = R(z \setminus x,z,y).$\\

Known prop\_deeac89a: $T(R(x,y,z),w) * (y * z) = (T(x,w) * y) * z.$\\

Known prop\_575d1ed9: $R(x,x \setminus e,y) = y / ((x \setminus e) * y).$\\

Known prop\_f338e359: $(e / x) * (x * y) = L(y,x,e / x).$\\

Known prop\_5a914e30: $R(x,y,y \setminus e) = (x * y) * (y \setminus e).$\\

Known prop\_cc8a9ae6: $R(x,e / y,y) = (x * (e / y)) * y.$\\

Known prop\_c2aa8580: $L(x \setminus (y \setminus z),x,y) = (y * x) \setminus z.$\\

Known prop\_b3890d2c: $R(x / y,y,y \setminus e) = x * (y \setminus e).$\\

Known prop\_f14899ed: $T(L(x / y,z,w),y) = L(y \setminus x,z,w).$\\

Known prop\_f2b7d0ab: $L(x \setminus T(y,z),x,z) = (z * x) \setminus (y * z).$\\

Known prop\_be6cad0a: $R((x / y) / z,z,y) = x / (z * y).$\\

Known prop\_1e558562: $L((x \setminus y) / z,z,x) = (z * ((x * z) \setminus y)) / z.$\\

Known prop\_e9eef609: $K(x \setminus e,x) = (x \setminus e) * x.$\\

Known prop\_9ee87fb5: $K(x,e / x) = x * (e / x).$\\

Known prop\_da958b3f: $L(x \setminus e,x,y) = (y * x) \setminus y.$\\

Known prop\_c3fa51e8: $R(e / x,x,y) = y / (x * y).$\\

Known prop\_ba6418d1: $R(e / x,x,y) \setminus y = x * y.$\\

Known prop\_d7dd57dd: $(x * y) \setminus ((z * x) * y) = R(T(z,x * y),x,y).$\\

Known prop\_e1aa92db: $(x * y) * K(y,x) = y * x.$\\

Known prop\_19fcac9b2: $R(x / y,z,w) * y = y * R(y \setminus x,z,w).$\\

Known prop\_3d75df700: $(x * y) * R((x * y) \setminus y,z,w) = (x * R(x \setminus e,z,w)) * y.$\\

Known prop\_acafcc6f0: $L(R(x \setminus (y \setminus z),w,u),x,y) = R((y * x) \setminus z,w,u).$\\

Known prop\_203fc9151: $x * (y * R(y \setminus (x \setminus y),z,w)) = (x * R(x \setminus e,z,w)) * y.$\\

Known prop\_2e844a2a9: $(x * y) * R((x * y) \setminus z,w,u) = x * (y * R(y \setminus (x \setminus z),w,u)).$\\

Known prop\_d9f457e09: $(x \setminus y) * R((x \setminus y) \setminus y,z,w) = R(x,z,w) * (x \setminus y).$\\

Known prop\_ce2987245: $x \setminus R(x,y,z) = R(x,y,z) * (x \setminus e).$\\

Known prop\_b7fe5fbfb: $K(x \setminus e,x) = (e / x) \setminus (x \setminus e).$\\

Known prov9\_7c96e347d4: $R(T(e / x,z),x,y) = T(y / (x * y),z).$\\

Known prov9\_3e047dc57d: $R(x,x \setminus e,y) \setminus y = (x \setminus e) * y.$\\

Known prov9\_ee78192c46: $R(x,y,x) * (y * x) = (y * x) * T(x,y).$\\

Known prov9\_062c221162: $T(x,y) = R(T(x,y * x),y,x).$\\

Known prov9\_d18167fcf7: $x * (T(x,y) \setminus e) = T(x,y) \setminus x.$\\

Known prov9\_6385279d78: $(x \setminus y) / (y / x) = (e / (y / x)) * (x \setminus y).$\\

Known prov9\_b192646899: $x * T(x \setminus e,y) = K(y \setminus (y / x),y).$\\

Known prov9\_2e3bc568bd\_alt1: $K(x \setminus (x / (e / y)),x) * y = T(y,x).$\\

Known prov9\_1ffb5e2572: $L(T(x \setminus e,z),x,y) = T((y * x) \setminus y,z).$\\

Known prov9\_7fed2c3e64: $(x * y) * T((x * y) \setminus x,z) = x * (y * T(y \setminus e,z)).$\\

Known prov9\_47e1e09ded: $(x * y) * T((x * y) \setminus y,z) = (x * T(x \setminus e,z)) * y.$\\

Known prov9\_a06014c62d\_com: $x * (x * T(x \setminus e,y)) = (x * T(x \setminus e,y)) * x.$\\

Known prov9\_a06014c62d: $T(x,x * T(x \setminus e,y)) = x.$\\

Known prov9\_49726cdcf0: $T(x,(x \setminus e) * x) = x.$\\

Known prov9\_a69214de59: $R(x,x \setminus e,x) = T(x,x \setminus e).$\\

Known prov9\_3a3c9a39ee: $T(x,x \setminus e) \setminus e = T(x \setminus e,x).$\\

Known prov9\_1cecad55d3: $T(x,e / x) = (x \setminus e) \setminus e.$\\

Known prov9\_13e5c8ed0a: $T(e / (e / x),x \setminus e) = x.$\\

Known prov9\_183b179b43: $K(x,x \setminus e) * x = T(x,x \setminus e).$\\

\end{multicols}
\section{Implementation details}
All models were implemented using PyTorch. All experiments for the Robinson and polynomial arithmetic were run on a 16 core Intel(R) Xeon(R) CPU E5-2670 0 @ 2.60GHz. The AIM experiments were run on a 72 core Intel(R) Xeon(R) Gold 6140 CPU @ 2.30GHz. All calculations were done on CPU.

The PPO implementation was adapted from \url{https://github.com/nikhilbarhate99/PPO-PyTorch}. The model was updated every 2000 timesteps, the PPO clip coefficient was set to 0.2. The number of epochs each update was set to 4. The learning rate was 0.002 and the discount factor $\gamma$ was set to 0.99. 

The ACER implementation was adapted from \url{https://github.com/dchetelat/acer}. The replay buffer size was 20,000. The truncation parameter was 10 and the model was updated every 100 steps. The replay ratio was set to 4. Trust region decay was set to 0.99 and the constraint was set to 1. The discount factor was set to 0.99 and the learning rate to 0.001. Off-policy minibatch size was set to 1.

The A2C and SIL implementations were based on Pytorch reinforcement learning actor-critic example code available at \url{https://github.com/pytorch/examples/tree/master/reinforcement_learning}. The code was modified to use the bootstrapped value estimate when doing on-policy updates. For SIL, we implemented the SIL loss on top of the A2C implementation. There is also a prioritized replay buffer with an exponent of 0.6, as in the original paper. Each epoch, 8000 (250 batches of size 32) transitions were taken from the prioritized replay buffer in the SIL step of the algorithm. The size of the prioritized replay buffer was 40,000. The critic loss weight was set to 0.01 as in the original paper.

For the 3SIL and behavioral cloning implementations, we sample 8000 transitions (250 batches of size 32) from the replay buffer or history. For the behavioral cloning, we used a buffer of size 40,000. The 3SIL history or replay buffer contains at a maximum \textit{k} solutions for each problem in the training set. When sampling transitions in 3SIL, we first sample a problem and then we sample a transition from the stored solutions for that problem in the buffer. 

For all approaches the environment was queried to find out which actions were available in the current state. The softmax outputs corresponding to invalid actions were set to 0 and the distribution normalized.

\section{Data processing and proving environment details}
For the polynomial arithmetic normalization dataset, there were a handful of problems that contained numbers larger than 100 in the normalized expression. These were filtered out as these caused large slowdowns due to the successor representation of numbers. 20 out of 6000 problems were filtered out by this procedure. 

AIMLEAP expects a distance estimate for each applicable action. This represents the estimated distance to a proof. This behavior was converted to a reinforcement learning setup by always setting the chosen action of the model to the minimum distance and all other actions to a distance larger than the maximum proof length. Only the chosen action is then carried out.

Versions of the automated theorem provers used: Version 2.5 of EProver (\url{https://wwwlehre.dhbw-stuttgart.de/~sschulz/E/E.html}), the Nov 2017 version of Prover9 (\url{https://github.com/ai4reason/Prover9}) and the Feb 2018 version of Waldmeister (\url{https://www.mpi-inf.mpg.de/departments/automation-of-logic/software/waldmeister/download}).
\clearpage
\section{Training of AIM models}
\begin{figure}[h!]
    \centering
    \includegraphics[width=0.6\textwidth]{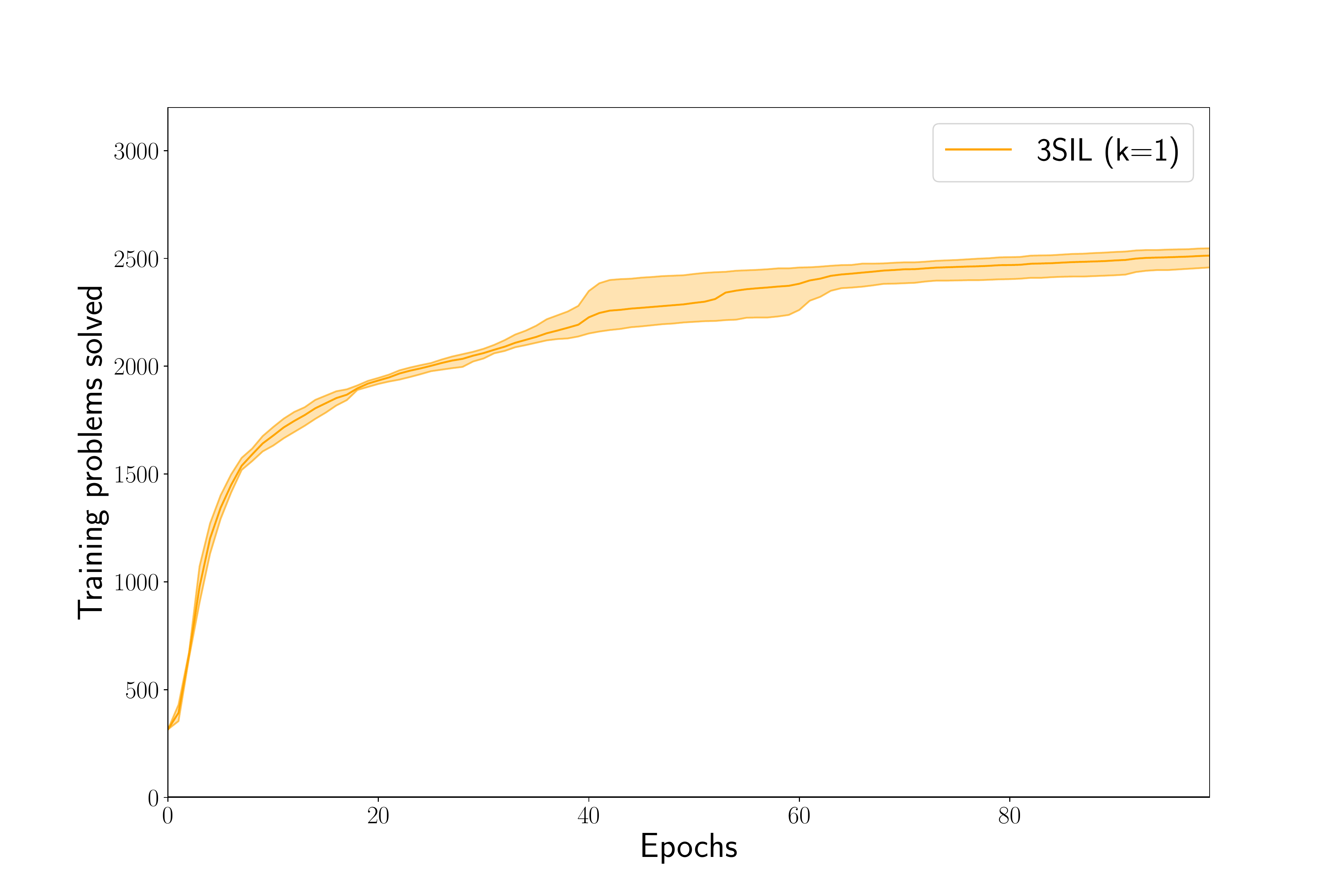}
    \caption{The number of training problems for which a solution was encountered and stored. At the start of the training, the model rapidly collects more solutions, but after 100 epochs, the process has slowed down and settled at about 2500 problems with known solutions. }
    \label{fig:my_label}
\end{figure}
\end{document}